\documentclass{article}
\usepackage{spconf,amsmath,graphicx}

\usepackage{times}
\usepackage{latexsym}

\usepackage[T1]{fontenc}
\usepackage{url}

\usepackage[utf8]{inputenc}

\usepackage{microtype}

\usepackage{graphicx}
\usepackage{comment}

\usepackage{microtype}
\usepackage{amsmath,amsfonts}

\newcommand{\ie}{\emph{i.e., }}

\title{VK-G2T: Vision and Context Knowledge enhanced Gloss2Text}
\name{Liqiang Jing$^{1}$, Xuemeng Song$^{1\star}$, Xinxing Zu$^2$, Na Zheng$^{3 }$, Zhongzhou Zhao$^{2}$, Liqiang Nie$^{4	}$
\thanks{$^\star$Xuemeng Song is the corresponding author.}
}
\address{$^{1}$Shandong University,  $^{2}$Alibaba Group,\\
$^{3 }$National University of Singapore, 
$^{4}$Harbin Institute of Technology (Shenzhen) 
}
\begin{document}
%
\maketitle
\begin{abstract}
Existing sign language translation methods follow a two-stage pipeline: first converting the sign language video to a gloss sequence (\ie Sign2Gloss) and then translating the generated gloss sequence into a spoken language sentence (\ie Gloss2Text). While previous studies have focused on boosting the performance of the Sign2Gloss stage, we emphasize the optimization of the Gloss2Text stage. However, this task is non-trivial due to two distinct features of Gloss2Text: (1) isolated gloss input and (2) low-capacity gloss vocabulary. To address these issues, we propose a vision and context knowledge enhanced Gloss2Text model, named VK-G2T, which leverages the visual content of the sign language video to learn the properties of the target sentence and exploit the context knowledge to facilitate the adaptive translation of gloss words. Extensive experiments conducted on a Chinese benchmark validate the superiority of our model.
\end{abstract}

\section{Introduction} \label{intro}
Sign language, consisting of a sequence of hand gestures instead of spoken words, is a vital communication tool for the deaf community to express their thoughts. 
However, understanding these hand gestures can be challenging for individuals without hearing impairments. Therefore, 
researchers have focused on the task of sign language translation (SLT) to bridge the communication gap between the deaf and the non-disabled by translating sign language videos into spoken language sentences. 
Combining the advanced deep learning technologies~\cite{liu2018attentive,liu2018cross,DBLP:conf/sigir/SongFHYLN18}, existing SLT methods~\cite{DBLP:conf/coling/YinR20,DBLP:conf/sitis/ArvanitisCK19} mainly adopt a two-stage pipeline consisting of: Sign2Gloss and  Gloss2Text, where the former stage works on converting the hand gesture sequence in the sign language video into a gloss sequence (\ie a sequence of isolated words with basic meaning, such as  ``I'', ``eat'', and ``good''), and the latter stage targets at translating the extracted gloss sequence into the coherent spoken sentence.


Previous SLT studies~\cite{DBLP:conf/coling/YinR20,DBLP:conf/iccv/MinHC021} focus on optimizing the first stage, \ie Sign2Gloss, while paying less attention to improving the performance of the second stage, \ie Gloss2Text. Specifically, existing SLT methods~\cite{DBLP:conf/cvpr/CamgozKHB20,phonenix,DBLP:journals/corr/smmtl} mainly directly adopt the existing neural machine translation models, such as Long Short-Term Memory~(LSTM)~\cite{DBLP:journals/neco/HochreiterS97}, Transformer~\cite{DBLP:conf/nips/VaswaniSPUJGKP17}, and the pretrained language model BART~\cite{DBLP:conf/acl/LewisLGGMLSZ20} to perform the task of Gloss2Text.

\begin{figure}
    \centering
    \includegraphics[scale=0.25]{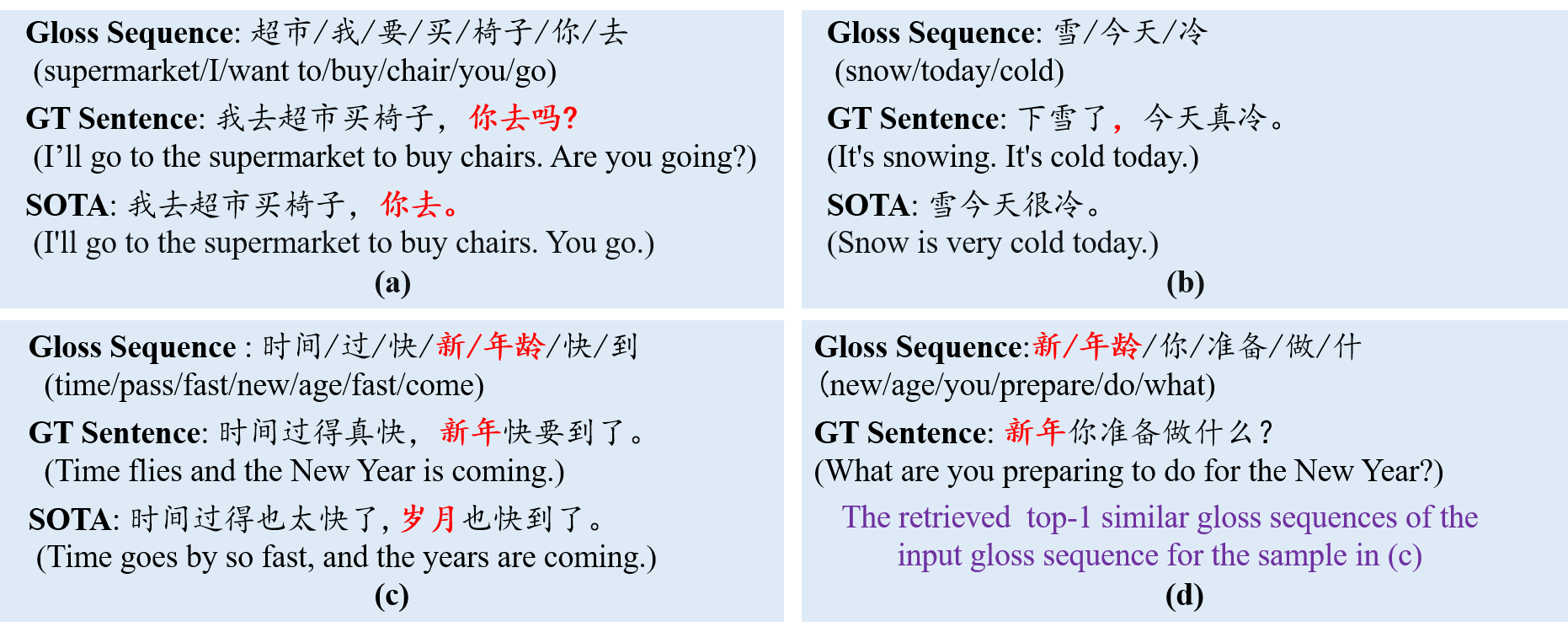}
    \caption{
    (a) Failure caused by incorrect sentence type prediction. (b) Failure caused by incorrect sentence structure prediction. (c) Failure caused by the low-capacity gloss vocabulary. (d) A retrieved sample most similar to that in (c). 
    GT: ground-truth. 
    The English texts are translated from the Chinese texts, where the Chinese texts with syntax errors are difficult to translate.}
    \label{fig:fig2}
\end{figure}

However, previous studies overlook the following two distinct features of the task of Gloss2Text. 1) Isolated gloss input. Unlike conventional machine translation models that process complete and natural language sentences,  Gloss2Text operates on sequences of isolated glosses without punctuation symbols. This distinction hinders the standard machine translation models from learning crucial sentence properties, such as sentence type (e.g., interrogative sentence) and sentence structure (e.g., whether the sentence has a pause), potentially resulting in incorrect translations. As can be seen from Figure~\ref{fig:fig2} (a) and (b), even the state-of-the-art SLT model, \ie SMMTL~\cite{DBLP:journals/corr/smmtl}, which directly employs mBART~\cite{DBLP:journals/tacl/LiuGGLEGLZ20} for Gloss2Text,  fails to predict the sentence property and hence generate the inappropriate translation,  due to the absence of punctuation symbols in the isolated gloss input. 
2) Low-capacity gloss vocabulary. 
The vocabulary of the sign language/gloss is pre-defined by experts, which is usually in a small scale and cannot cover all the spoken language words. 
As a result, the input low-capacity gloss is difficult to be interpreted by conventional machine translation models. As can be seen from Figure~\ref{fig:fig2} (c), the state-of-the-art SLT model SMMTL fails to adaptively translate ``new/age'' into the spoken language phrase ``New Year'' based on the given gloss context.

Intuitively, the sentence type can be conveyed by non-hand movements~\cite{jushi} and the sentence structure can be reflected by gestural pause~\cite{tingdun}. Therefore, to address the challenge of isolated gloss input, we propose to incorporate the visual content to compensate for the information loss in the isolated gloss input, and hence benefit the property prediction of the target spoken sentence and improve the generation of spoken language. Meanwhile, to cope with the low capacity issue, we propose to leverage the context knowledge in the dataset to facilitate the adaptive Gloss2Text translation. Specifically, for a given gloss sequence, we consider similar gloss sequences from the training set along with their corresponding ground truth sentences as the context knowledge. As shown in Figure \ref{fig:fig2} (d), referring to the context knowledge could guide the model to translate ``new/age'' to ``New Year'', rectifying the failure sample in Figure \ref{fig:fig2} (c). Accordingly, in this work, we propose to explore the visual content of the original sign language video as well as the context knowledge in the dataset to enhance the Gloss2Text performance. Towards this end, we propose a vision and context knowledge enhanced Gloss2Text model, named VK-G2T, which consists of three components: Vision-based Sentence Property Learning, Context Knowledge-enhanced Gloss Sequence Embedding, and Spoken Language Generation. 

\begin{figure}
    \centering
    \includegraphics[width=0.5\textwidth]{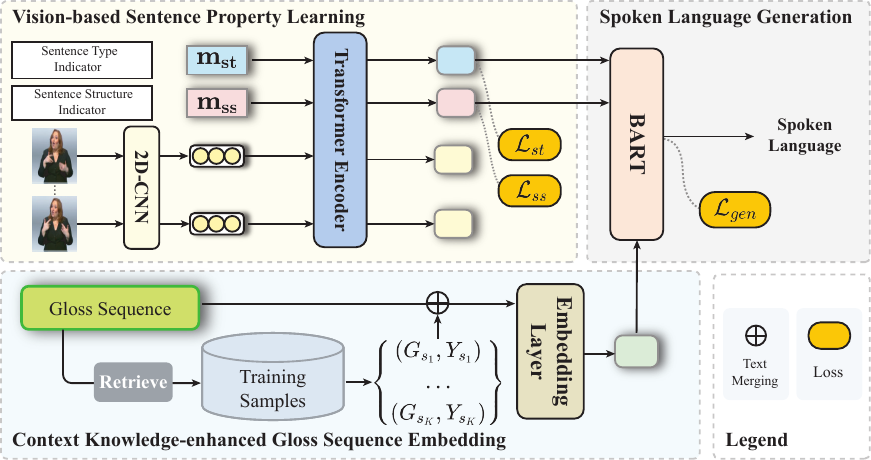}
    \caption{Illustration of the proposed scheme VK-G2T, which consists of three components: Vision-based Sentence Property Learning, Context Knowledge-enhanced Gloss Sequence Embedding, and Spoken Language Generation.}
    \label{fig:method}
\end{figure}

\section{Method}
Following the standard SLT setting~\cite{phonenix,csl-daily}, we suppose that there is a training set $\mathcal{D}$, which consists of a set of $N$ training samples, \ie $\mathcal{D}=\{(V_i, G_i, Y_i)\}_{i=1}^N$. 
$V_i=(v^1_i,v^2_i,\cdots,v^{N_{V_i}})$ denotes the $i$-th sign language video, where $v^j_i$ is the $j$-th frame of $V_i$ and $N_{V_i}$ is the total number of frames. 
$G_i = (g^1_i,g^2_i,\cdots,g^{N_{G_i}}_i)$ is the gloss sequence for the video $V_i$ with $N_{G_i}$ gloss words. 
$Y_i=(y^1_i,y^2_i,\cdots,y^{N_{Y_i}}_i)$ is the target spoken language sentence with $N_{Y_i}$ words for the gloss sequence $G_i$ or the video $V_i$. 
Then our goal is to learn a model $\mathcal{F}$ that can translate a gloss sequence into the spoken language sentence by additionally referring to the original sign language video as follows,
\begin{equation}
    \hat{Y_i} = \mathcal{F}({G}_i, V_i|\Theta),
\end{equation}
where $\Theta$ refers to the set of to-be-learned parameters of the model $\mathcal{F}$. $\hat{Y_i}$ is the generated spoken language sentence. 
Notably, we temporally omit the subscript $i$ for simplicity.
Next, we give the details of the three components of our proposed VK-G2T model, as shown in Figure~\ref{fig:method}. 

\subsection{Vision-based Sentence Property Learning}
In this component, we aim to learn the property of the target sentence based on the original sign language video, to 
compensate the information loss during the process of Sign2Gloss. In particular, we consider two sentence properties that largely affect the quality of the generated spoken language: 1) sentence type and 2) sentence structure. 
As Transformer has 
achieved remarkable success on various sequence modeling 
tasks~\cite{DBLP:conf/acl/MaL0C22,DBLP:journals/corr/smmtl}, we resort it to derive the sentence type and sentence structure information from the given sign language video. Specifically, 
we first resort to the pretrained 2D-CNN neural network~\cite{DBLP:journals/tmm/CuiLZ19} following previous studies on SLT~\cite{phonenix,DBLP:journals/corr/smmtl} to extract the visual feature of each frame in the video as follows, 
\begin{equation}
    \mathbf{f}_j = \mathbf{2D\_CNN}(v^j),
\end{equation}
where $\mathbf{f}_j \in \mathbb{R}^{D_1}$ is the feature of the $j$-th frame of the video $V$ and $D_1$ is the feature dimension. 

Thereafter, to model the sentence type and sentence structure information from the given video, we introduce two prefix tokens $\mathbf{m}_{st} \in \mathbb{R}^{D_1}$ and $\mathbf{m}_{ss} \in \mathbb{R}^{D_1}$ as the indicators to encode the sentence type and sentence structure information, respectively. We then feed the concatenation of $[\mathbf{m}_{st}, \mathbf{m}_{ss}]$ and the visual feature sequence into a Transformer encoder as follows,
\begin{equation} \label{constrast_prj}
 \left\{
 \begin{aligned}
  &\mathbf{Z}=  Transformer(\mathbf{F}+\mathbf{E}_{pos}),\\
  & \mathbf{F} = [\mathbf{m}_{st};\mathbf{m}_{ss};\mathbf{f}_1;\cdots;\mathbf{f}_{N_V}],
  \end{aligned}
 \right.
 \end{equation}
where $[;]$ denotes concatenation operation, 
and $\mathbf{F}\in \mathbb{R}^{(N_V+2) \times D_1}$ is the concatenated feature embedding sequence. 
 $\mathbf{E}_{pos} \in \mathbb{R}^{(N_v+2) \times D_1}$ is the positional embedding for modeling the positional information within the input sequence, which is pre-defined based on the sine and cosine function~\cite{DBLP:conf/nips/VaswaniSPUJGKP17,bert}. $\mathbf{Z} \in \mathbb{R}^{(N_v+2) \times D_1}$ is the output of the Transformer encoder, whose first two columns correspond to the final embeddings of the two prefix indicator tokens. Formally, we define $\mathbf{z}_{st}=\mathbf{Z}[1,:]$ and $\mathbf{z}_{ss}=\mathbf{Z}[2,:]$ to denote the final indicator embeddings for the sentence type and sentence structure, respectively.

\textbf{Sentence Type Learning.} 
To enhance the sentence type indicator embedding learning, we employ a linear neural network to classify the sentence types and adopt the cross-entropy loss function for supervision. 
In this work, we classify the sentences into two most common types: interrogative and non-interrogative. 
Mathematically, we have
\begin{equation} \label{task1}
 \left\{
 \begin{aligned}
  &\mathbf{\hat{p}}_{st} = \operatorname{softmax}(\textbf{W}_1 \textbf{z}_{st} +\textbf{b}_1),\\
  &\mathcal{L}_{st} = - \sum_{i=1}^{2} p_{st}^i \log \hat{p}_{st}^i,
  \end{aligned}
 \right.
 \end{equation}
where $\mathbf{W}_1 \in \mathbb{R}^{2 \times D_1}$, $\mathbf{b}_1 \in \mathbb{R}^{2}$ are the trainable parameters. 
$\mathbf{\hat{p}}_{st}=[ \hat{p}_{st}^1, \hat{p}_{st}^2]\in \mathbb{R}^2 $ is the predicted sentence pattern probability distribution. 
$\mathbf{p}_{st}=[ {p}_{st}^1, {p}_{st}^2]\in \mathbb{R}^2$ is the ground truth, which can be obtained by checking whether the target spoken sentence $Y$ contains the question mark ``?''.
Specifically, if the target spoken sentence is an interrogative, $\mathbf{p}_{st}=[1,0]$, otherwise, $\mathbf{p}_{st}=[0,1]$.

\textbf{Sentence Structure Learning}. 
Similar to sentence type learning, in this part, we adopt another linear neural network to classify whether the sentence has a pause and then use a cross-entropy loss function for supervision. We also consider the most common cases for sentence structure classification: compound and non-compound sentences, \ie we consider whether the sentence has a pause/sub-sentences. Similar to sentence type learning, we have the predicted sentence structure probability distribution $\mathbf{\hat{p}}_{ss}=[\hat{p}_{ss}^1,\hat{p}_{ss}^2]\in \mathbb{R}^2$ and corresponding loss function $\mathcal{L}_{ss}$.  
The corresponding ground truth equals $[1,0]$ if the target spoken language sentence $Y$ contains the pause indicator (\ie the comma ``,'') and $[0,1]$ otherwise.


\subsection{Context Knowledge-enhanced Gloss Sequence Embedding}
As aforementioned, the vocabulary of the sign language is pre-defined by experts and in limited size, which cannot cover all the spoken language words. 
For example, in the vocabulary of Chinese sign language, there is a gloss word corresponding to the word of ``age'', but no gloss word for the  word of ``year''. 
Hence, one gloss word should be flexibly translated into different spoken words under different contexts. For instance, the gloss word ``age'' should be translated to ``Year'' when the context is ``new/age'', while ``age'' when the context is ``old/age''. 
Towards this end, we resort to retrieving the similar gloss sequences to the given gloss sequence from the whole training dataset, and taking into account their target sentences as the context knowledge to enhance the semantic understanding of the gloss sequence as well as the spoken language translation.

To be specific, we first retrieve the top $K$ similar gloss sequences from the training set for the given gloss sequence ${G}$ based on the similarity metric BM25 score. Let $G_{s_k}$ be the $k$-th similar gloss sequence for $G$ in the training set, where  $s_k\in \{1,\cdots, N\}$. Then we incorporate the retrieved $K$ similar gloss sequences, along with their corresponding target spoken language sentences as context knowledge to the input gloss sequence $G$ as follows,

\begin{equation} \label{eq9}
    G' = \{{G},(G_{s_1},Y_{s_1}),\cdots,(G_{s_K},Y_{s_K})\},
\end{equation}
where $Y_{s_k}$ is the ground truth sentence for the gloss sequence $G_{s_k}$. 

Next, we use the embedding layer of BART~\cite{DBLP:conf/acl/LewisLGGMLSZ20} which has shown powerful ability in multimodal generation tasks~\cite{ml1,ml2,ml3,ml4}, to get the embedding of $G'$ as follows,
\begin{equation}
    \mathbf{E}_g=BART\_Embed (G'),
\end{equation}
where $\mathbf{E}_g=[\textbf{e}_1, \textbf{e}_2,\cdots,     \textbf{e}_U]$ denotes the context knowledge-enhanced gloss sequence embedding. $U$ is the total number of tokens in ${G}'$.

 \subsection{Spoken Language Generation}
After obtaining the sentence type embedding $\mathbf{z}_{st}$, sentence structure embedding $\mathbf{z}_{ss}$, and auxiliary knowledge-enhanced gloss embedding $\mathbf{E}_g$, we can proceed with the target sentence generation. Here we utilize BART, which has demonstrated strong generation ability, as the generator backbone. Specifically, we first concatenate $\mathbf{z}_{st}$, $\mathbf{z}_{ss}$ and $\mathbf{E}_g$ along with the positional embedding into the BART encoder 
as follows,
\begin{equation} \label{eq11}
 \left\{
 \begin{aligned}
    &\mathbf{Z}_g=BART\_{Encoder}(\mathbf{E}+\mathbf{E'}_{pos}), \\
    &\textbf{E} =[\mathbf{z}_{st};\mathbf{z}_{ss};\mathbf{E}_g],
  \end{aligned}
 \right.
 \end{equation}
where  $\mathbf{E'}_{pos} \in \mathbb{R}^{(U+2) \times D_1}$ is the predefined positional embedding, and $\mathbf{Z}_g \in \mathbb{R}^{(U+2) \times D_1}$ is the encoded context representation. 

We then feed the encoded representation into the autoregressive decoder of BART, which generates the next token by referring to all the previously decoded outputs as follows,
\begin{equation}
\hat{\textbf{y}}_i=BART\_{Decoder}(\mathbf{Z}_g, Y_{<i}),
\end{equation}
where $i \in [1,N_Y]$ and $\hat{\textbf{y}}_i \in \mathbb{R}^{|\mathcal{V}|}$ is the predicted for the $i$-th token probability distribution of the target sentence. $|\mathcal{V}|$ is the size of the vocabulary.  $Y_{<i}$ is the the previous $i-1$ tokens in the target spoken sentence.

For optimization, we introduce the standard cross-entropy loss function as follows,
\begin{equation} \label{prob}
  \mathcal{L}_{gen} = -1/N_Y\sum_{i=1}^{N_Y}\log (\hat{\textbf{y}}_i[t]),
 \end{equation}
where $\hat{\textbf{y}}_i[t]$ is the element of  $\hat{\textbf{y}}_i$ that corresponds to the $i$-th token of the ground-truth target sentence, and $N_Y$ is the total number of tokens in the target sentence $Y$.
\subsection{Training}
Ultimately, we combine all the loss functions for optimizing our VK-G2T as $\mathcal{L} = \mathcal{L}_{gen} + \alpha \mathcal{L}_{st}+ \beta \mathcal{L}_{ss} \label{eq16}$. 
where $\alpha$ and $\beta$ are the non-negative hyperparameters used for balancing the effect of each component on the entire training process. 

\section{Experiment}

\begin{table}[!t]
    \centering
    \caption{Performance comparison on CSL-Daily.  
    }
    \label{tab:rq1}
    \resizebox{0.5\textwidth}{!}{
    \begin{tabular}{l|ccccc} 
    \hline
    &\multicolumn{5}{c}{TEST}\\ 
         Model &  ROUGE-L & BLEU-1 & BLEU-2 & BLEU-3&BLEU-4\\ \hline
         Seq2Seq~\cite{phonenix}& 
         50.04&53.70&37.04& 25.80& 18.51 \\ 
         Transformer~\cite{DBLP:conf/cvpr/CamgozKHB20}  & 39.67& 41.74 & 26.71 & 16.90 & 10.71 \\
         
        TIN-SLT & 41.98 &51.30 &31.95 &21.81 &15.51 \\ 
         SMMTL~\cite{DBLP:journals/corr/smmtl} & {56.61} & {59.83} & {45.18} &{34.54} & {26.75} \\ 
         SMMTL-cBART & \underline{62.93} & \underline{64.49} & \underline{51.88} &\underline{42.05} & \underline{34.40}\\ \hline
         VK-G2T & \textbf{65.51} & \textbf{67.16} & \textbf{54.54}& \textbf{44.49} & \textbf{36.83}\\  
        \hline

    \end{tabular}}
\end{table}



\begin{table}[!t]
    \centering
    \caption{Ablation study of our model on CSL-Daily.  }
    \label{tab:ablation}
    \resizebox{0.5\textwidth}{!}{
    \begin{tabular}{l|ccccc} 
    \hline
    & \multicolumn{5}{c}{TEST}\\ 
         Model &  ROUGE-L & BLEU-1 & BLEU-2 & BLEU-3&BLEU-4\\ \hline
             w/o-Property & 63.78& 65.05 & 52.64& 42.92 & 35.04 \\

            w/o-Type  & 64.77&65.92 & {53.51}&{43.78} & {36.28}   \\ 

         w/o-Stru &  64.70& 65.85 & 53.24& 43.50&36.04 \\ 
        w/o-PropertyObj &  63.67& 64.95 & 52.71& 42.90&35.11 \\

         w/o-ContKnow &  {64.98}&{66.02}&53.47&43.71&36.21 \\ \hline

         VK-G2T & \textbf{65.51} & \textbf{67.16} & \textbf{54.54}& \textbf{44.49} & \textbf{36.83}\\ \hline
    \end{tabular}}
\end{table}




\subsection{Experimental Setup}
For evaluation, we selected public datasets: {CSL-Daily}~\cite{csl-daily}. 
Due to the language difference, we used the pretrained Chinese BART model\footnote{\url{https://huggingface.co/fnlp/bart-base-chinese}.} to initialize the generator of our model. The hyper-parameters $\alpha$ and $\beta$ are both set to $1$, while $D_1$ is set to $768$. The Transformer encoder used for vision-based sentence property learning consists of $2$ layers and $8$ attention heads. 
The length of the target sentence $|T|$ is unified to $60$ and the total training epoch is set to $40$. 
We adopted the Adam optimizer~\cite{adam} with a learning rate of $10e-5$. The batch size is $8$. For each gloss sequence, we retrieved $3$ similar gloss sequences ($K=3$).
For evaluation, similar to previous work~\cite{DBLP:journals/corr/smmtl,csl-daily,phonenix}, {we adopted BLEU-1 to BLEU-4~\cite{DBLP:conf/acl/PapineniRWZ02}} and ROUGE-L~\cite{lin-2004-rouge} as evaluation metrics. 

\subsection{Model Comparison}

Table~\ref{tab:rq1} presents a performance comparison between our model and all baselines on the {CSL-Daily} dataset~\cite{csl-daily}. 
From this table, we make the following observations. 1) Our model surpasses all the baselines for all the metrics, which demonstrates the superiority of our proposed model.
2) SMMTL, which employs a generative pretrained language model as the generator backbone, surpasses all baselines that are trained from scratch (\ie Seq2Seq, Transformer, and TIN-SLT) across all metrics. 
This indicates that leveraging the linguistic knowledge acquired by the pretrained language model from massive texts can promote the Gloss2Text performance. 3) Our VK-G2T exceeds SMMTL-cBART and SMMTL across different metrics on the {CSL-Daily} dataset. This suggests the benefits of mining the sentence property from the video content as well as the context knowledge of the gloss sequence.

\subsection{Ablation Study}
To assess the significance of each component in our model, we conducted experiments on the following variations. 
1) \textbf{w/o-Property}. To justify the sentence property learning component, we removed both sentence property indicators and their corresponding losses by setting $ \textbf{E} = \mathbf{E}_g$, and $\alpha=\beta=0$ in Eqn.(\ref{eq16}). 
2) \textbf{w/o-Type}. To explore the effect of sentence type learning, we removed the sentence type indicator and its corresponding loss. 
3) \textbf{w/o-Stru}. Similarly, to show the effect of sentence structure learning, we removed the sentence structure indicator and its corresponding loss. 
 4)  \textbf{w/o-PropertyObj}. To determine the impact of vision-based property learning originates from the learning objective design rather than simple vision content incorporation, we designed this derivative by still inputting the visual modality and deriving the two indicators but setting $\alpha=\beta=0$ in Eqn.(\ref{eq16}).  
5) \textbf{w/o-ContKnow}. To demonstrate the necessity of incorporating the context knowledge, we removed the context knowledge by setting ${G}' = \{G\}$ in Eqn.(\ref{eq9}). 
 
We presented the ablation study results of our model VK-G2T on both the development set and testing set of CSL-Daily in Table~\ref{tab:ablation}. From this table, we noticed that our model outperforms all the derivatives, highlighting the benefits of sentence property learning and context knowledge incorporation in sign language translation. In particular, 
we observed that w/o-PropertyObj has a similar performance to w/o-Property, which discards both sentence property indicators and hence only considers the gloss sequence input without taking the visual input into account. This indicates that without the explicit supervision signal for sentence property learning, even incorporating the visual modality in w/o-PropertyObj, its benefit to sign language translation is very limited.


\section{Conclusion and Future Work}
In this paper, we present a vision and context knowledge enhanced Gloss2Text model,  which leverages the visual modality to learn the properties of the target sentence and incorporates the context knowledge to enhance the adaptive translation of the gloss words. 
Extensive experiments on a public dataset demonstrate the superiority of our model over existing cutting-edge methods and verify the effectiveness of each devised module. 
In the future, we plan to expand the evaluation of our model to datasets in more languages. 

\section{Acknowledgement}
This work was supported in part by the National Natural Science Foundation of China under Grant 62376137, and in part by the Shandong Provincial Natural Science Foundation under Grant ZR2022YQ59.

\bibliographystyle{IEEEbib}
\bibliography{strings,refs}

\begin{thebibliography}{10}

\bibitem{liu2018attentive}
Meng Liu, Xiang Wang, Liqiang Nie, Xiangnan He, Baoquan Chen, and Tat-Seng
  Chua,
\newblock ``Attentive moment retrieval in videos,''
\newblock in {\em SIGIR}, 2018.

\bibitem{liu2018cross}
Meng Liu, Xiang Wang, Liqiang Nie, Qi~Tian, Baoquan Chen, and Tat-Seng Chua,
\newblock ``Cross-modal moment localization in videos,''
\newblock in {\em Multimedia}, 2018.

\bibitem{DBLP:conf/sigir/SongFHYLN18}
Xuemeng Song, Fuli Feng, Xianjing Han, Xin Yang, Wei Liu, and Liqiang Nie,
\newblock ``Neural compatibility modeling with attentive knowledge
  distillation,''
\newblock in {\em {SIGIR}}, 2018.

\bibitem{DBLP:conf/coling/YinR20}
Kayo Yin and Jesse Read,
\newblock ``Better sign language translation with stmc-transformer,''
\newblock in {\em COLING}, 2020.

\bibitem{DBLP:conf/sitis/ArvanitisCK19}
Nikolaos Arvanitis, Constantinos Constantinopoulos, and Dimitrios~I.
  Kosmopoulos,
\newblock ``Translation of sign language glosses to text using
  sequence-to-sequence attention models,''
\newblock in {\em SITIS}, 2019.

\bibitem{DBLP:conf/iccv/MinHC021}
Yuecong Min, Aiming Hao, Xiujuan Chai, and Xilin Chen,
\newblock ``Visual alignment constraint for continuous sign language
  recognition,''
\newblock in {\em ICCV}, 2021.

\bibitem{DBLP:conf/cvpr/CamgozKHB20}
Necati~Cihan Camg{\"{o}}z, Oscar Koller, Simon Hadfield, and Richard Bowden,
\newblock ``Sign language transformers: Joint end-to-end sign language
  recognition and translation,''
\newblock in {\em CVPR}, 2020.

\bibitem{phonenix}
Necati~Cihan Camg{\"{o}}z, Simon Hadfield, Oscar Koller, Hermann Ney, and
  Richard Bowden,
\newblock ``Neural sign language translation,''
\newblock in {\em CVPR}, 2018.

\bibitem{DBLP:journals/corr/smmtl}
Yutong Chen, Fangyun Wei, Xiao Sun, Zhirong Wu, and Stephen Lin,
\newblock ``A simple multi-modality transfer learning baseline for sign
  language translation,''
\newblock in {\em CVPR}, 2022.

\bibitem{DBLP:journals/neco/HochreiterS97}
Sepp Hochreiter and J{\"{u}}rgen Schmidhuber,
\newblock ``Long short-term memory,''
\newblock {\em Neural Computation}, 1997.

\bibitem{DBLP:conf/nips/VaswaniSPUJGKP17}
Ashish Vaswani, Noam Shazeer, Niki Parmar, Jakob Uszkoreit, Llion Jones,
  Aidan~N. Gomez, Lukasz Kaiser, and Illia Polosukhin,
\newblock ``Attention is all you need,''
\newblock in {\em NeurIPS}, 2017.

\bibitem{DBLP:conf/acl/LewisLGGMLSZ20}
Mike Lewis, Yinhan Liu, Naman Goyal, Marjan Ghazvininejad, Abdelrahman Mohamed,
  Omer Levy, Veselin Stoyanov, and Luke Zettlemoyer,
\newblock ``{BART:} denoising sequence-to-sequence pre-training for natural
  language generation, translation, and comprehension,''
\newblock in {\em ACL}, 2020.

\bibitem{DBLP:journals/tacl/LiuGGLEGLZ20}
Yinhan Liu, Jiatao Gu, Naman Goyal, Xian Li, Sergey Edunov, Marjan
  Ghazvininejad, Mike Lewis, and Luke Zettlemoyer,
\newblock ``Multilingual denoising pre-training for neural machine
  translation,''
\newblock {\em TACL}, 2020.

\bibitem{jushi}
Junhui Yang,
\newblock ``An introduction to csl/chinese bilingual education,''
\newblock {\em Chinese Journal of Special Education}, 2002.

\bibitem{tingdun}
Qing Liu,
\newblock ``A study on non-manual feature in natural sign language of deaf
  people,''
\newblock {\em Cultural and Educational Materials}, 2014.

\bibitem{csl-daily}
Hao Zhou, Wengang Zhou, Weizhen Qi, Junfu Pu, and Houqiang Li,
\newblock ``Improving sign language translation with monolingual data by sign
  back-translation,''
\newblock in {\em CVPR}, 2021.

\bibitem{DBLP:conf/acl/MaL0C22}
Zhiyuan Ma, Jianjun Li, Guohui Li, and Yongjing Cheng,
\newblock ``Unitranser: {A} unified transformer semantic representation
  framework for multimodal task-oriented dialog system,''
\newblock in {\em ACL}, 2022.

\bibitem{DBLP:journals/tmm/CuiLZ19}
Runpeng Cui, Hu~Liu, and Changshui Zhang,
\newblock ``A deep neural framework for continuous sign language recognition by
  iterative training,''
\newblock {\em TMM}, 2019.

\bibitem{bert}
Jacob Devlin, Ming{-}Wei Chang, Kenton Lee, and Kristina Toutanova,
\newblock ``{BERT:} pre-training of deep bidirectional transformers for
  language understanding,''
\newblock in {\em NAACL}, 2019.

\bibitem{ml1}
Liqiang Jing, Yiren Li, Junhao Xu, Yongcan Yu, Pei Shen, and Xuemeng Song,
\newblock ``Vision enhanced generative pre-trained language model for
  multimodal sentence summarization,''
\newblock {\em Mach. Intell. Res.}, 2023.

\bibitem{ml2}
Dengtian Lin, Liqiang Jing, Xuemeng Song, Meng Liu, Teng Sun, and Liqiang Nie,
\newblock ``Adapting generative pretrained language model for open-domain
  multimodal sentence summarization,''
\newblock in {\em {SIGIR}}, 2023.

\bibitem{ml3}
Liqiang Jing, Xuemeng Song, Kun Ouyang, Mengzhao Jia, and Liqiang Nie,
\newblock ``Multi-source semantic graph-based multimodal sarcasm explanation
  generation,''
\newblock in {\em ACL}, 2023.

\bibitem{ml4}
Liqiang Jing, Xuemeng Song, Xuming Lin, Zhongzhou Zhao, Wei Zhou, and Liqiang
  Nie,
\newblock ``Stylized data-to-text generation: {A} case study in the e-commerce
  domain,''
\newblock {\em {ACM} Trans. Inf. Syst.}, 2024.

\bibitem{adam}
Diederik~P. Kingma and Jimmy Ba,
\newblock ``Adam: {A} method for stochastic optimization,''
\newblock in {\em ICLR}, 2015.

\bibitem{DBLP:conf/acl/PapineniRWZ02}
Kishore Papineni, Salim Roukos, Todd Ward, and Wei{-}Jing Zhu,
\newblock ``Bleu: a method for automatic evaluation of machine translation,''
\newblock in {\em ACL}, 2002.

\bibitem{lin-2004-rouge}
Chin-Yew Lin,
\newblock ``{ROUGE}: A package for automatic evaluation of summaries,''
\newblock in {\em Text Summarization Branches Out}, 2004.

\end{thebibliography}

\end{document}